\documentclass[letterpaper]{article} 
\usepackage{aaai24}  
\usepackage{times}  
\usepackage{helvet}  
\usepackage{courier}  
\usepackage[hyphens]{url}  
\usepackage{graphicx} 
\usepackage{natbib}  
\usepackage{caption} 
\usepackage{parskip}

\usepackage{algorithm}
\usepackage{algorithmic}
\usepackage{multirow}
\usepackage{booktabs}
\usepackage{amsmath}
\usepackage[table,xcdraw]{xcolor}
\usepackage{placeins}
\usepackage{subcaption}

\newcommand{\ms}[2]{{#1}{\footnotesize $\,\pm${#2}}} 
\newcommand{\msb}[2]{{\textbf{#1}} $\,\pm${\footnotesize {#2}}} 
\newcommand{\ourmethod}{TCBC}
\newcommand{\vx}{\boldsymbol{x}}
\newcommand{\vy}{y}
\newcommand{\vu}{\boldsymbol{u}}
\newcommand{\pr}{p}

\newcommand{\soft}{\operatorname{softmax}}
\newcommand{\para}{\theta}
\usepackage{newfloat}
\usepackage{listings}
\DeclareCaptionStyle{ruled}{labelfont=normalfont,labelsep=colon,strut=off} 
\floatstyle{ruled}
\newfloat{listing}{tb}{lst}{}
\floatname{listing}{Listing}
\title{Twice Class Bias Correction for Imbalanced Semi-Supervised Learning}
\author{
    Lan Li,
    Bowen Tao,
    Lu Han,
    De-chuan Zhan,
    Han-jia Ye\thanks{Corresponding author.}
}
\affiliations{
    National Key Laboratory for Novel Software Technology, Nanjing University\\
    School of Artificial Intelligence, Nanjing University\\
    Nanjing, 210023, China

    \{lil, taobw, hanlu, yehj\}@lamda.nju.edu.cn, zhandc@nju.edu.cn
}

\usepackage{bibentry}

\begin{document}

\maketitle

\begin{abstract}
Differing from traditional semi-supervised learning, class-imbalanced semi-supervised learning presents two distinct challenges: (1) The imbalanced distribution of training samples leads to model bias towards certain classes, and (2) the distribution of unlabeled samples is unknown and potentially distinct from that of labeled samples, which further contributes to class bias in the pseudo-labels during training. To address these dual challenges, we introduce a novel approach called \textbf{T}wice \textbf{C}lass \textbf{B}ias \textbf{C}orrection (\textbf{TCBC}). We begin by utilizing an estimate of the class distribution from the participating training samples to correct the model, enabling it to learn the posterior probabilities of samples under a class-balanced prior. This correction serves to alleviate the inherent class bias of the model. Building upon this foundation, we further estimate the class bias of the current model parameters during the training process. We apply a secondary correction to the model's pseudo-labels for unlabeled samples, aiming to make the assignment of pseudo-labels across different classes of unlabeled samples as equitable as possible. Through extensive experimentation on CIFAR10/100-LT, STL10-LT, and the sizable long-tailed dataset SUN397, we provide conclusive evidence that our proposed TCBC method reliably enhances the performance of class-imbalanced semi-supervised learning. 
\end{abstract}

\section{Introduction}
Semi-supervised learning (SSL) \cite{van2020survey,li2019safe,yang2019semi,yang2019comprehensive} has shown promise in using unlabeled data to reduce the cost of creating labeled data and improve model performance on a large scale. In SSL, many algorithms generate pseudo-labels \cite{lee2013pseudo}  for unlabeled data based on model predictions, which are then utilized to regularize model training. However, most of these methods assume that the data is balanced across classes. In reality, many real-world datasets exhibit imbalanced distributions \cite{buda2018systematic,byrd2019effect,pouyanfar2018dynamic, li2022aligning}, with some classes being much more prevalent than others. This imbalance affects both the labeled and unlabeled samples, resulting in biased pseudo-labels that further worsen the class imbalance during training and ultimately hinder model performance. Recent research \cite{wei2021crest,lee2021abc,guo2022class, wei2022transfer,Tao2023CLAFCL} has highlighted the significant impact of class imbalance on the effectiveness of pseudo-labeling methods. Therefore, it is crucial to develop SSL algorithms that can effectively handle class imbalance in both labeled and unlabeled data, leading to improved performance in real-world scenarios.

Unlike traditional SSL techniques that assume identical distributions of labeled and unlabeled data, this paper addresses a more generalized scenario of imbalanced SSL. Specifically, we consider situations where the distribution of unlabeled samples is unknown and may diverge from the distribution of labeled samples~\cite{oh2022daso, l2ac, wei2023towards}.

\begin{figure*}
     \centering
     \begin{subfigure}[b]{0.33\textwidth}
         \centering
         \includegraphics[width=\textwidth]{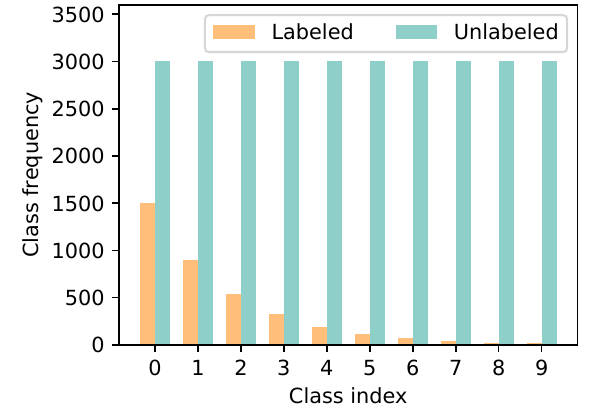}
         \caption{Class frequency}
         \label{fig:1a}
     \end{subfigure}
     \hfill
     \begin{subfigure}[b]{0.33\textwidth}
         \centering
         \includegraphics[width=\textwidth]{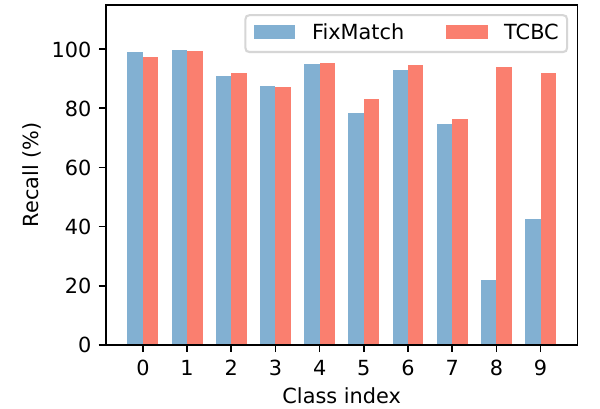}
         \caption{Recall on test set}
         \label{fig:1b}
     \end{subfigure}
     \hfill
     \begin{subfigure}[b]{0.33\textwidth}
         \centering
         \includegraphics[width=\textwidth]{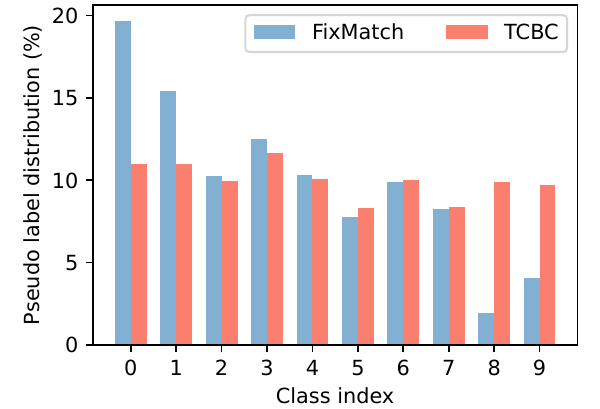}
         \caption{pseudo-label distribution}
         \label{fig:1c}
     \end{subfigure}
        \caption{Experiments on CIFAR-10. (a) Class distribution of the labeled and unlabeled data.
(b) Recall for each class generated from the test set. (c) The pseudo-label distribution 
generated from the unlabeled set.}
        \label{fig:three graphs}
\end{figure*}
In this context, two challenges need to be addressed: (1) How to mitigate the model's class bias induced by training on imbalanced data, and (2) How to leverage the model's predictions on unlabeled samples during the training process to obtain improved pseudo-labels. To elucidate these challenges, we devised an experiment, as depicted in Figure~\ref{fig:1a}, where labeled samples follow a long-tailed distribution, and unlabeled samples follow a uniform distribution. Figure~\ref{fig:1b} illustrates the recall performance of the FixMatch model trained under this scenario on the test set. Despite the presence of numerous minority class samples in the unlabeled data, due to the class imbalance in the labeled samples, the eventual model still exhibits significant class bias. Figure~\ref{fig:1c} represents the pseudo-label class distribution on unlabeled data obtained by Fixmatch, highlighting a notable class bias present in Fixmatch's pseudo-labels, which can adversely affect model training. To address these two challenges within imbalance SSL, we introduce a novel approach termed ``Twice Class Bias Correction" (TCBC).

The primary challenge of the first issue lies in the potential inconsistency between the class distributions of labeled and unlabeled samples. During training, the class distribution of training samples may undergo substantial fluctuations, rendering it infeasible to rely on the assumption of consistent class distribution to reduce model bias. To address this challenge, we dynamically estimate the class distribution of participating training samples. Leveraging the assumption of consistent class-conditional probabilities, we guide the model to learn a reduced class bias objective on both labeled and pseudo-labeled samples, specifically targeting the posterior probabilities of samples under a class-balanced prior. As depicted in Figure~\ref{fig:1b}, our approach significantly diminishes the model's bias across different classes.

The complexity of the second challenge lies in the presence of class bias in the model during training and the unknown distribution of unlabeled samples. This results in uncontrollable pseudo-labels acquired by the model on unlabeled data. A balanced compromise solution is to ensure that the model acquires pseudo-labels as equitably as possible across different classes. Hence, we introduce a method based on the model's output on samples to estimate the model's class bias under current parameters. We leverage this bias to refine predictions on unlabeled samples, thereby reducing class bias in pseudo-labels. As illustrated in Figure~\ref{fig:1c}, our approach achieves a less biased pseudo-label distribution on class-balanced unlabeled samples.

Our primary contributions are as follows:
(1) We present a novel technique that harnesses the class distribution of training samples to rectify the biases introduced by class imbalance in the model's learning objectives. (2) We introduce a method to evaluate the model's class bias under the current model parameter conditions during the training process and utilize it to refine pseudo-labels.
(3) Our approach is straightforward yet effective, as demonstrated by extensive experiments in various imbalanced SSL settings, highlighting the superiority of our method.
Code and appendix is publicly available at https://github.com/Lain810/TCBC.

\section{Related Work}
\textbf{Class Imbalanced learning} attempts to learn models that generalize well on each class from imbalanced data.  Resampling and reweighting are two commonly used methods. Resampling methods balance the number of training samples for each class in the training set by undersampling ~\cite{buda2018systematic,drummond2003c4} the majority classes or oversampling~\cite{buda2018systematic,byrd2019effect,pouyanfar2018dynamic} the minority classes. Reweighting~\cite{cui2019class,huang2016learning,wang2017learning} methods assign different losses to different training samples of each class or each example. 
In addition, some works have used logits compensation \cite{cao2019learning,menon2020long,ren2020balanced,ye2020identifying} based on class distribution or transfer learning~\cite{yin2019feature,chu2020feature,ye2021procrustean} to address this problem.

\noindent\textbf{Semi-supervised learning} try to improve the model’s performance by leveraging unlabeled data~\cite{berthelot2019mixmatch}. A common approach in SSL is to utilize model predictions to generate pseudo-labels for each unlabeled data and use these pseudo-labels for supervised training. Recent SSL algorithms, exemplified by FixMatch~\cite{fixmatch}, achieve enhanced performance by encouraging consistent predictions between two different views of an image and employing consistency regularization. While these methods have seen success, most of them are based on the assumption that labeled and unlabeled data follow a uniform label distribution. When applied to class-imbalanced scenarios, the performance of these methods can significantly deteriorate due to both model bias and pseudo-label bias.

\begin{figure*}[t!]
\centering
\includegraphics[width=0.8\textwidth]{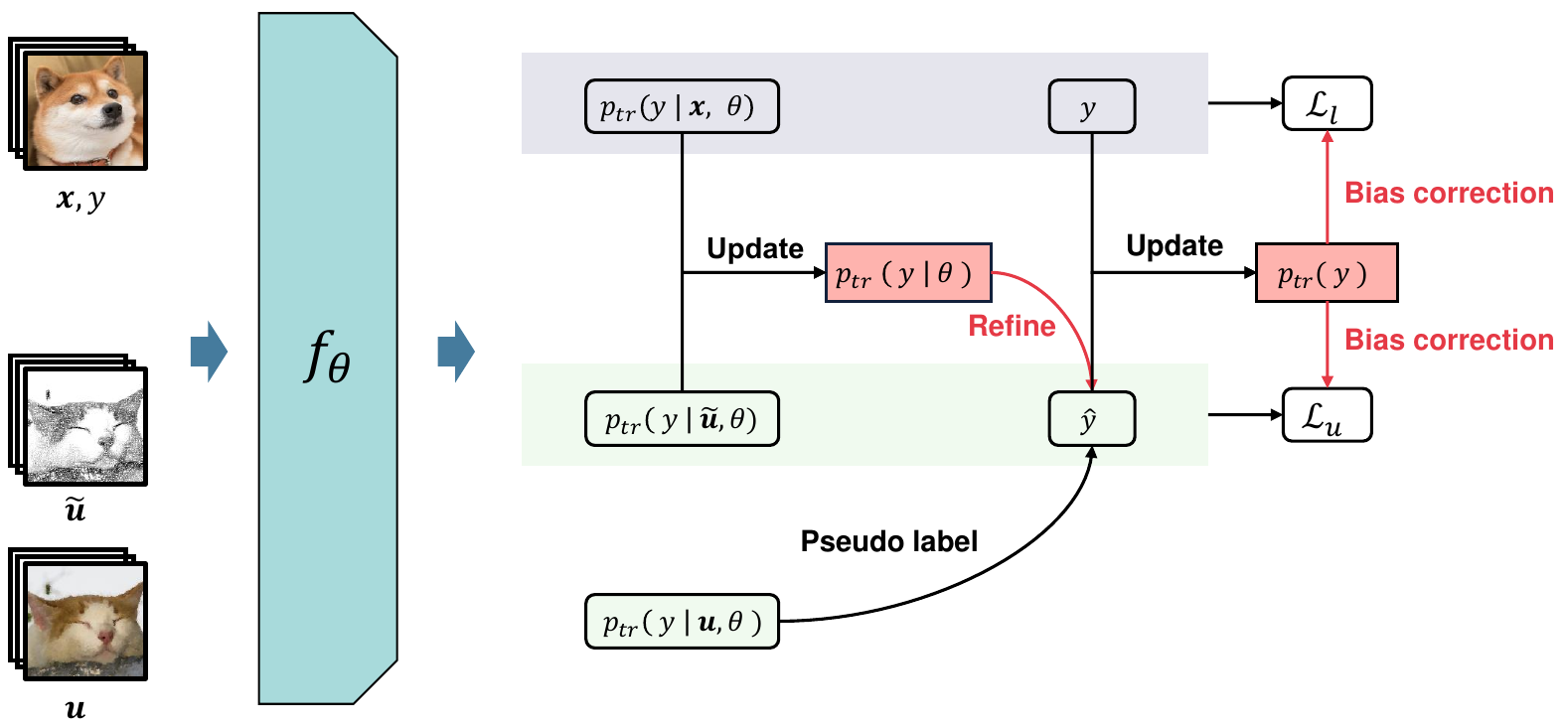}
\caption{Illustration of the proposed method. Our approach corrects the class bias in the model's learning objectives by estimating the class distribution based on the participating training data. Simultaneously, it refines pseudo-labels by estimating the model's class bias under the current parameters, reducing bias in the pseudo-labels.}
\label{fig2}
\end{figure*}
\noindent\textbf{Class imbalanced semi-supervised learning} has garnered widespread attention due to its alignment with real-world tasks. The DARP~\cite{darp} refines initial pseudo-labels through convex optimization, aiming to alleviate distribution bias resulting from imbalanced and unlabeled training data. In contrast, CREST~\cite{wei2021crest} employs a combination of re-balancing and distribution alignment techniques to mitigate training bias. ABC~\cite{lee2021abc} introduces an auxiliary balanced classifier trained through down-sampling of majority classes to enhance generalization.
However, many existing methods assume a similarity between marginal distributions of labeled and unlabeled data classes, an assumption that often doesn't hold or remains unknown before training. To address this limitation, DASO~\cite{oh2022daso} combines pseudo-labels from both linear and similarity-based classifiers, leveraging their complementary properties to combat bias. L2AC~\cite{l2ac} introduces a bias adaptive classifier to tackle the issue of training bias in imbalanced semi-supervised learning tasks.

\section{Methodology}
\subsection{Preliminaries}
Assuming the existence of a labeled set denoted by $\mathcal{D}_l = \{(\vx_n, \vy_n)\}_{n=1}^N$ and an unlabeled set denoted by $\mathcal{D}_u = \{\vu_m\}_{m=1}^M$, where $\vx_n, \vu_m \in \mathcal{X}$ represents training samples in the input space, and $\vy_n \in \mathcal{Y}$ represents the labels assigned to labeled samples, with $\mathcal{Y}$ denoting the label space.  The class distribution of labeled data and unlabeled data is denoted by~$\pr_{l}(\vy)$ and $\pr_{u}(\vy)$, respectively. Moreover, we denote $N_k$ and $M_k$ as the number of labeled and unlabeled samples in class $k$, respectively. Without loss of generality, we assume that the classes are arranged in descending order based on the number of training samples, such that $N_1 \geq N_2 \geq \ldots \geq N_K$. The goal of imbalanced SSL is to learn the model $f$ that generalizes well on each class from imbalanced data, parameterized by $\theta$.

In SSL, one effective method involves utilizing pseudo-labeling techniques to enhance the training dataset with pseudo-labels for unlabeled data. In pseudo-labeling SSL, each unlabeled sample is provided with a pseudo-label based on the model's prediction. An optimization problem with an objective function $\mathcal{L} = \mathcal{L}_s+\lambda \mathcal{L}_u$ is utilized to train the model on both labeled and pseudo-labeled samples. The loss function consists of two terms: the supervised loss $\mathcal{L}_s$ computed on labeled data and the unsupervised loss $\mathcal{L}_u$ computed on unlabeled data. The parameter $\lambda$ is used to balance the loss from labeled data and pseudo-labeled data. The computation formula for the supervised loss on labeled samples is given by
\begin{equation}
    \mathcal{L}_{s}=\frac{1}{N} \sum_{x_i\in B_l } \mathcal{H}\left(\vy_{i}, p\left(\vy | \vx_{i}\right)\right),
\end{equation}
where $B_l$ denotes a iteration of labeled data sampled from $D_l$. $\pr \left(\vy | \vx_{i}\right)=\mathrm{softmax}(f(\vx_{i}))$ represents the output probability, and $\mathcal{H}$ is the cross-entropy loss. Similarly, the loss function on unlabeled samples can also be formulated as
\begin{equation}
\mathcal{L}_{u}=\frac{1}{M} \sum_{\vu_j\in B_{u}}  \mathcal{M} \cdot \mathcal{H}\left(\hat{y}_j, p\left(\vy | \tilde{\vu}_j\right)\right),
\end{equation}
where $\mathcal{M} = \mathbf{I}\left[\max \left(\hat{p}\left(y | \vu_j \right)\right) \geq \tau_{c}\right]$ and $\mathbf{I}$ is the indicator function, $\tau_{c}$ is the threshold.
$\hat{y}_j = \arg\max \hat{p}\left(y | \vu_j \right)$ denotes the pseudo-label assigned by the model to the weakly augmented sample $\vu_j$, and $\tilde{\vu}_j$ represents the unlabeled samples that undergo strong augmentation.
       
\subsection{Model bias correction}

In imbalanced SSL, the first challenge to address is \textit{how to learn a model that is devoid of class bias}. Generally, achieving a evaluation  model without class bias involves approximating a Bayesian optimal model under a class-balanced distribution. This entails minimizing the loss on data where the marginal distribution follows a uniform distribution, $\pr_{ev}(y)=1/K$. The corresponding posterior probability of the samples is denoted as $\pr_{ev}(\vy| \vx)$.

When directly optimizing surrogate loss, such as the softmax cross-entropy loss, on training data with an imbalanced class distribution $\pr_{tr}(\vy)$, the learned posterior probability of the samples becomes $\pr_{tr}(\vy|\vx)$, which differs from $\pr_{ev}(\vy| \vx)$  and tends to favor the classes with more samples. Therefore, resolving this challenge involves addressing the mismatch between $\pr_{tr}(\vy|\vx)$ and $\pr_{ev}(\vy|\vx)$.

When considering imbalanced SSL, the situation becomes more complex. The training data consist of labeled data (with a marginal distribution $\pr_{l}(\vy)$) and unlabeled data (with a marginal distribution $\pr_{u}(\vy)$). Here, $\pr_{l}(\vy)$ represents a known long-tail distribution, while $\pr_{u}(\vy)$ represents an unknown distribution that may differ from $\pr_{l}(\vy)$.

Assuming that the training samples, including labeled and unlabeled samples, are drawn from the same probability distribution of the corresponding class,
\begin{equation}
    \pr_{l}(\vx|\vy)=\pr_{u}(\vx|\vy)=\pr(\vx|\vy)  \quad \vx \in \mathcal{X}, \vy \in \mathcal{Y}. \label{eq:post}
\end{equation}
By applying the Bayes' theorem, we establish that $\pr(\vy|\vx) \propto \pr(\vx|\vy) \pr(\vy)$. If we consider labeled and unlabeled losses separately, due to the fact that $\pr_{l}(\vy) \neq \pr_{u}(\vy)$, the posterior probabilities learned from these two parts are different. Specifically, let's assume a sample $\vx_i$ exists in both the labeled and unlabeled data. The posterior probabilities $\pr_{tr}(\vy | \vx_i)$ learned by the model represent a weighted average of $\pr_{l}(\vy | \vx_i)$ and $\pr_{u}(\vy | \vx_i)$. We denote the marginal probability distribution corresponding to $\pr_{tr}(\vy | \vx)$ as $\pr_{tr}(\vy)$. According to Eq.~\eqref{eq:post}, and considering $\pr_{ev}(y) = 1/K$, we have:
\begin{equation}
    \begin{aligned} \pr_{tr}(\vy | \vx)  & \propto \pr( \vx | \vy) \cdot \pr_{tr}({y})  \\ & \propto \pr_{ev}(\vy | \vx) / \pr_{ev}(\vy) \cdot \pr_{tr}({y}) \\ 
    & \propto \pr_{ev}(\vy | \vx)  \cdot \pr_{tr}({y}).
    \end{aligned} \label{eq:evtra}
\end{equation}
Because we aim to learn $\pr_{ev}(\vy|\vx)$, which can be represented as $\pr_{ev}(\vy|\vx)=\soft(f(\vx))$, but what we learn through the loss $\mathcal{L}$ is $\pr_{tr}(\vy|\vx)$. According to Equation \eqref{eq:evtra}, we have:
\begin{equation}
    \begin{aligned}
     \pr_{tr}(\vy|\vx) & \propto \pr_{ev}(\vy | \vx)  \cdot \pr_{tr}({y}) \\
      & \propto \soft(f(\vx)+\ln{\pr_{tr}(\vy)}).
    \end{aligned} \label{eq:evtra2}
\end{equation}
Therefore, the $\pr(\vy|\vx_i)$ in $\mathcal{L}_s$ and $\pr(\vy|\tilde{\vu}_j)$ in $\mathcal{L}_u$ should be:
\begin{equation}
\begin{aligned}
    \pr_{tr}(\vy|\vx_i) &= \soft(f(\vx)+\ln{\pr_{tr}(\vy)}) \\
    \pr_{tr}(\vy|\tilde{\vu}_j)&= \soft(f(\tilde{\vu}_j)+\ln{\pr_{tr}(\vy)}).
\end{aligned}
\end{equation}
This loss can be considered as an extension of the logit adjustment loss~\cite{menon2020long}~or balance softmax~\cite{ren2020balanced}~in imbalanced SSL. As the class distribution of unlabeled data is unknown and varies during training, we need to estimate $\pr_{tr}({y})$. We use the class distribution of the participating training samples in the most recent $T$ iterations as an estimate of $\pr_{tr}({y})$. Additionally, since there are weights associated with the loss in $\mathcal{L}_u$, we treat it as a form of undersampling. Consequently, the number of samples for class $y$ in a batch is given by:
\begin{equation}
    \operatorname{count}(\vy)=\sum_{ \vx_i \in B_l}\mathbf{I}(\vy_i=\vy)+\sum_{ \vu_j \in B_u}\lambda\mathcal{M}\cdot \mathbf{I}(\hat{y}_j=\vy). \nonumber
\end{equation}
In the experiments, we set $T$ to be $50*K$. 

\begin{figure}
     \centering
     \begin{subfigure}[b]{0.23\textwidth}
         \centering
         \includegraphics[width=\textwidth]{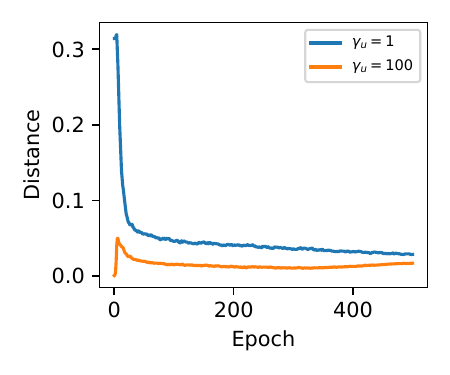}
         \caption{Model bias correction}
         \label{fig:2a}
     \end{subfigure}
     \hfill
     \begin{subfigure}[b]{0.23\textwidth}
         \centering
         \includegraphics[width=\textwidth]{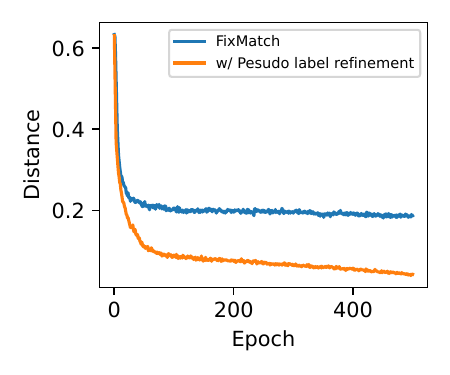}
         \caption{Pseudo-label refinement}
         \label{fig:2b}
     \end{subfigure}
        \caption{(a) L2 Distance between $p_{tr}(y)$ and ture distrustion in consistent($\gamma_u=100$) and inconsistent($\gamma_u=1$) setting when applying model bias correction. (b) L2 distance between the pseudo-label distribution and the unlabeled distribution with pseudo-label refinement.}
        \label{fig:three graphs}
\end{figure}

To investigate the correlation between $\pr_{tr}(y)$ and the true distribution (representing the class distribution of $\mathcal{D}_l\cup\mathcal{D}_u$), we monitored the L2 distance between $\pr_{tr}(y)$ and the true distribution during the training process. Figure~\ref{fig:2a} shows the results for both cases: when the distributions of labeled and unlabeled samples are consistent and when they are inconsistent. It can be observed that when the distributions are consistent, $\pr_{tr}(y)$ maintains a small distance from the true distribution. When the distributions are inconsistent, $\pr_{tr}(y)$ continuously approaches the true distribution.

\subsection{Pseudo-label refinement}
In the previous section, we utilized an estimate of the marginal distribution $\pr_{tr}(y)$ of the training data to learn a class-balanced model. However, \textit{is it optimal to use this model to generate pseudo-labeled data for training during the training process?}

We conducted a parametric decomposition of Equations \eqref{eq:evtra} and \eqref{eq:evtra2}, taking into further consideration the parameters $\theta_t$ of the model at the $t$-th iteration:
\begin{equation}
    \begin{aligned} \pr_{bal}(\vy | \vx ; \para_{t}) & \propto \pr( \vx | \vy ; \para_t) \cdot \pr_{tr}(\vy | \para_t) / \pr_{tr}(\vy | \para_t) \\ & \propto \pr_{tr}(\vy | \vx ; \para_t) / p_{tr}(y | \theta_t) \\ & \propto \operatorname{softmax}\left(f(\vx)+\ln\pr_{tr}(\vy)-\ln \pr_{tr}(y | \para_t)\right),
    \end{aligned} \nonumber
\end{equation}
where $\pr_{bal}(\vy | \vx ; \para_{t})$ represents the posterior probability under the uniform distribution prior for a given model parameters $\para_{t}$. It takes into account the model parameters more explicitly than $\pr_{ev}(\vy | \vx)$.
To compute $\pr_{{bal}}(\vy | \vx ; \para_{t})$, a crucial step is to estimate the class prior $p_{tr}(y | \theta_t)$ under the current parameters. Referring to \citet{hong2023long}, we can estimate $p_{tr}(y | \theta_t)$ through the model's outputs on the training samples:
\begin{equation}
    \begin{aligned}
    \pr_{tr}(\vy | \theta_t) &= \mathbf{E}_{\vx_i \in \mathcal{D}_{tr}} \pr_{tr}(\vy | \vx ; \para_t) \\
    &= \mathbf{E}_{\vx_i \in \mathcal{D}_{tr}} \left(\pr_{ev}(\vy|\vx; \para_t) \cdot \pr_{tr}(\vy)\right),
    \end{aligned} \nonumber
\end{equation}
where $\mathcal{D}_{tr}$ represents the dataset composed of participating training samples. Let $d_y(\para_t) = \ln\pr_{tr}(\vy) - \ln p_{tr}(y | \para_t)$, we have:
\begin{equation}
    \begin{aligned}
    d_y(\para_t) &= \ln\pr_{tr}(\vy)-\ln p_{tr}(\vy | \para_t) \\
    &= \ln\left(\frac{\pr_{tr}(\vy)}{\mathbf{E}_{\vx_i \in \mathcal{D}_{tr}} (\soft\left(f(\vx) + \ln\pr_{tr}(\vy)\right))}\right) \\
    &=-\ln\mathbf{E}_{\vx_i \in \mathcal{D}_{tr}}\left( \frac{e^{f^{\vy}(\vx_i)}}{\sum_{k=1}^K e^{f^{k}(\vx_i)+\ln \pr_{tr}(k)}} \right).
    \end{aligned} \nonumber
\end{equation}

\begin{table*}[t]
    \centering
    \caption{Comparison of accuracy (\%) on CIFAR10-LT and under $\gamma_l=\gamma_u$ and $\gamma_l\neq\gamma_u$ setting. We set $\gamma_l$ to $100$ and $150$ for CIFAR10-LT. We use \textbf{bold} to mark the best results.}
    \label{}
    \resizebox{\linewidth}{!}{%
\begin{tabular}{@{}l|cccc|cccc@{}}
\toprule
                                     & \multicolumn{4}{c|}{CIFAR10-LT}                                                                                            & \multicolumn{4}{c}{CIFAR10-LT}                                                                                            \\ \cmidrule(l){2-9} 
                                     & \multicolumn{2}{c|}{$N_1=1500,M_1=3000$}                                         & \multicolumn{2}{c|}{$N_1=500,M_1=4500$} & \multicolumn{2}{c|}{$N_1=1500,M_1=3000$}                                         & \multicolumn{2}{c}{$N_1=500,M_1=4500$} \\ \cmidrule(l){2-9} 
                                     & $\gamma_l=100$   & \multicolumn{1}{c|}{$\gamma_l=150$}                           & $\gamma_l=100$     & $\gamma_l=150$     & $\gamma_l=100$   & \multicolumn{1}{c|}{$\gamma_l=100$}                           & $\gamma_l=100$     & $\gamma_l=100$    \\
\multirow{-4}{*}{Algorithm}          & $\gamma_u=100$   & \multicolumn{1}{c|}{$\gamma_u=150$}                           & $\gamma_u=100$     & $\gamma_u=150$     & $\gamma_u=1$     & \multicolumn{1}{c|}{$\gamma_u=1/100$}                         & $\gamma_u=1$       & $\gamma_u=1/100$  \\ \midrule
Supervised                           & \ms{61.9}{0.41}  & \multicolumn{1}{c|}{\ms{58.2}{0.29}}                          & \ms{47.3}{0.95}    & \ms{44.2}{0.33}    & \ms{61.9}{0.41}  & \multicolumn{1}{c|}{\ms{61.9}{0.41}}                          & \ms{47.3}{0.95}    & \ms{47.3}{0.95}   \\
FixMatch                             & \ms{77.5}{1.32}  & \multicolumn{1}{c|}{\ms{72.4}{1.03}}                          & \ms{67.8}{1.13}    & \ms{62.9}{0.36}    & \ms{81.5}{1.15}  & \multicolumn{1}{c|}{\ms{71.8}{1.70}}                          & \ms{73.0}{3.81}    & \ms{62.5}{0.49}   \\
w~/~DARP                              & \ms{77.8}{0.63}  & \multicolumn{1}{c|}{\ms{73.6}{0.73}}                          & \ms{74.5}{0.78}    & \ms{67.2}{0.32}    & \ms{84.6}{0.34}  & \multicolumn{1}{c|}{\ms{80.0}{0.93}}                          & \ms{82.5}{0.75}    & \ms{70.1}{0.22}   \\
w~/~CReST+                             & \ms{78.1}{0.42}  & \multicolumn{1}{c|}{\ms{73.7}{0.34}}                          & \ms{76.3}{0.86}    & \ms{67.5}{0.45}    & \ms{86.4}{0.42}  & \multicolumn{1}{c|}{\ms{72.9}{2.00}}                          & \ms{82.2}{1.53}    & \ms{62.9}{1.39}   \\
w~/~ABC                                & \ms{83.8}{0.36}  & \multicolumn{1}{c|}{\ms{80.1}{0.45}}                          & \ms{78.9}{0.82}    & \ms{66.5}{0.78}    & \ms{82.8}{0.61}  & \multicolumn{1}{c|}{\ms{77.9}{0.86}}                          & \ms{73.1}{0.33}    & \ms{66.6}{0.39}   \\
w~/~DASO                              & \ms{79.1}{0.75}  & \multicolumn{1}{c|}{\ms{75.1}{0.77}}                          & \ms{76.0}{0.37}    & \ms{70.1}{1.81}    & \ms{88.8}{0.59}  & \multicolumn{1}{c|}{\ms{80.3}{0.65}}                          & \ms{86.6}{0.84}    & \ms{71.0}{0.95}   \\
w~/~L2AC                               & \ms{82.1}{0.57}  & \multicolumn{1}{c|}{\ms{77.6}{0.53}}                          & \ms{76.1}{0.45}    & \ms{70.2}{0.63}    & \ms{89.5}{0.18}  & \multicolumn{1}{c|}{\ms{82.2}{1.23}}                          & \ms{87.1}{0.23}    & \ms{73.1}{0.28}   \\
\rowcolor[HTML]{C0C0C0} 
\cellcolor[HTML]{C0C0C0}w~/~\ourmethod~(ours) & \msb{84.0}{0.55} & \multicolumn{1}{c|}{\cellcolor[HTML]{C0C0C0}\msb{80.4}{0.58}} & \msb{80.3}{0.45}   & \msb{75.2}{0.32}   & \msb{92.8}{0.42} & \multicolumn{1}{c|}{\cellcolor[HTML]{C0C0C0}\msb{85.7}{0.17}} & \msb{92.4}{0.29}  & \msb{79.9}{0.41} \\ \bottomrule
\end{tabular}
    } \label{table:c10}
\end{table*}
\begin{table*}[t]
    \centering
    \caption{Comparison of accuracy (\%) on CIFAR100-LT and STL10 under $\gamma_l=\gamma_u$ and $\gamma_l\neq\gamma_u$ setting. We set $\gamma_l$ to $10$ and $20$  for them. We use \textbf{bold} to mark the best results.}
    \label{}
    \resizebox{\linewidth}{!}{%
\begin{tabular}{@{}l|cccc|cccc@{}}
\toprule
                                     & \multicolumn{4}{c|}{CIFAR100-LT}                                                                                        & \multicolumn{4}{c}{STL10}                                                                                               \\ \cmidrule(l){2-9} 
                                     & \multicolumn{4}{c|}{$N_1=150,M_1=300$}                                                                                  & \multicolumn{2}{c|}{$N_1=150,M_1=100k$}                                        & \multicolumn{2}{c}{$N_1=450,M_1=100k$} \\ \cmidrule(l){2-9} 
                                     & $\gamma_l=10$    & \multicolumn{1}{c|}{$\gamma_l=20$}                            & $\gamma_l=10$     & $\gamma_l=10$    & $\gamma_l=10$   & \multicolumn{1}{c|}{$\gamma_l=20$}                           & $\gamma_l=10$      & $\gamma_l=20$     \\
\multirow{-4}{*}{Algorithm}          & $\gamma_u=10$    & \multicolumn{1}{c|}{$\gamma_u=20$}                            & $\gamma_u=1$      & $\gamma_u=1/10$  & $\gamma_u=N/A$  & \multicolumn{1}{c|}{$\gamma_u=N/A$}                          & $\gamma_u=N/A$     & $\gamma_u=N/A$    \\ \midrule
Supervised                           & \ms{46.9}{0.22}  & \multicolumn{1}{c|}{\ms{41.2}{0.15}}                          & \ms{46.9}{0.22}   & \ms{46.9}{0.22}  & \ms{40.9}{4.11} & \multicolumn{1}{c|}{\ms{36.4}{3.12}}                         & \ms{60.1}{5.22}    & \ms{50.6}{4.15}   \\
FixMatch                             & \ms{46.5}{0.06}  & \multicolumn{1}{c|}{\ms{50.7}{0.25}}                          & \ms{58.6}{0.73}   & \ms{57.6}{0.53}  & \ms{51.6}{2.32} & \multicolumn{1}{c|}{\ms{47.6}{4.87}}                         & \ms{72.4}{0.71}    & \ms{64.0}{2.27}   \\
w~/~DARP                              & \ms{58.1}{0.44}  & \multicolumn{1}{c|}{\ms{52.2}{0.66}}                          & \ms{55.8}{0.30}   & \ms{51.6}{0.83}  & \ms{66.9}{1.66} & \multicolumn{1}{c|}{\ms{59.9}{2.17}}                         & \ms{75.6}{0.45}    & \ms{72.3}{0.60}   \\
w~/~CReST+                             & \ms{57.4}{0.18}  & \multicolumn{1}{c|}{\ms{52.1}{0.21}}                          & \ms{59.3}{0.88}   & \ms{56.6}{0.23}  & \ms{61.2}{1.27} & \multicolumn{1}{c|}{\ms{57.1}{3.67}}                         & \ms{71.5}{0.96}    & \ms{68.5}{1.88}   \\
w~/~ABC                                & \ms{59.1}{0.21}  & \multicolumn{1}{c|}{\ms{53.7}{0.55}}                          & \ms{61.3}{0.53}   & \ms{59.8}{0.63}  & \ms{00.6}{0.53} & \multicolumn{1}{c|}{\ms{00.6}{0.53}}                         & \ms{78.6}{0.86}    & \ms{66.9}{0.98}   \\
w~/~DASO                               & \ms{59.2}{0.35}  & \multicolumn{1}{c|}{\ms{52.9}{0.42}}                          & \ms{59.8}{0.32}   & \ms{59.7}{0.43}  & \ms{70.1}{1.19} & \multicolumn{1}{c|}{\ms{65.7}{1.78}}                         & \ms{78.4}{0.80}    & \ms{75.3}{0.44}   \\
w~/~L2AC                               & \ms{57.8}{0.19}  & \multicolumn{1}{c|}{\ms{52.6}{0.13}}                          & \ms{61.3}{0.44}   & \ms{59.2}{0.36}  & \ms{75.1}{0.44} & \multicolumn{1}{c|}{\ms{72.3}{0.22}}                         & \ms{79.9}{0.52}    & \ms{77.0}{0.65}   \\
\rowcolor[HTML]{C0C0C0} 
\cellcolor[HTML]{C0C0C0}w~/~\ourmethod~(ours) & \msb{59.4}{0.28} & \multicolumn{1}{c|}{\cellcolor[HTML]{C0C0C0}\msb{53.9}{0.72}} & \msb{63.2}{0.65} & \msb{59.9}{0.42} & \msb{77.6}{0.93} & \multicolumn{1}{c|}{\cellcolor[HTML]{C0C0C0}\msb{74.9}{1.42}} & \msb{84.5}{0.52}    & \msb{82.6}{1.23}   \\ \bottomrule
\end{tabular}
    } \label{table:c100}
\end{table*}
Utilizing $d_y(\para_t)$ to modify $f(\vx)$ allows us to acquire pseudo-labels under the current parameters that mitigate class bias. However, an accurate estimation of $d_y(\para_t)$ necessitates considering the entire dataset. To ensure stable and efficient estimation of $d_y(\para_t)$, we devised a momentum mechanism that leverages expectations computed over each iteration for momentum updates:

\begin{equation}
    d_y(\para_{t+1})  = m \cdot d_y(\para_{t}) + (1-m) \cdot d'_y(\para_{t}), \nonumber
\end{equation}
where $m\in [0, 1)$ is a momentum coefficient,  $d'_y(\para_{t})$ is computed utilizing data from the $t$-th iteration. Consequently, the refined pseudo-label can be expressed as follows:
$$\hat{y}^{re}_{j} = \arg\max \left( \soft\left(f\left(\vx_i\right)+d_y\left(\para_{t+1}\right)\right) \right).$$
Fixing $\pr_{tr}(y)$ at $1/K$ leads the algorithm to degenerate into a process that exclusively incorporates pseudo-label refinement into the FixMatch.	We conducted a comparative analysis between the approach that exclusively employs pseudo-label refinement and the original FixMatch, aiming to explore the characteristics of pseudo-label refinement. Experimental trials were carried out utilizing unlabeled samples distributed uniformly, aligning with the conditions depicted in Figure~\ref{fig:1a}. Figure~\ref{fig:2b} visualizes the L2 distance between the distribution of pseudo-labels and the uniform distribution.	Clearly, with the advancement of training, the refined distribution of pseudo-labels gradually converges toward the uniform distribution.	Our proposed approach successfully mitigates the class bias present in pseudo-labels.	

In summary, Figure~\ref{fig2} presents the overall training procedure for TCBC. The labeled and unlabeled loss are given by:
$$\mathcal{L}_{s}=\frac{1}{N} \sum_{x_i\in B_l } \mathcal{H}\left(\vy_{i}, \pr_{tr}\left(\vy | \vx_{i}\right)\right),$$
$$  \mathcal{L}_{u}=\frac{1}{M} \sum_{\vu_j\in B_{u}}  \mathcal{M} \cdot \mathcal{H}\left(\hat{y}^{re}_{j}, \pr_{tr}\left(\vy | \tilde{\vu}_j\right)\right).$$
\section{Experiments}

This section presents a comprehensive evaluation of our algorithm's performance within the context of imbalanced SSL in classification problems.

\subsection{Experimental setup} 
\label{sec:setup}
\textbf{Datasets} We conduct experiments on three benchmarks including CIFAR10, CIFAR100 \cite{cifar100} and STL10 \cite{coates2011analysis}, which are commonly used in imbalance learning and SSL task. Results on real-world dataset, SUN-397, are also given in appendix. To validate the effectiveness of \ourmethod, we evaluate TCBC under various ratio of class imbalance. For imbalance types, we adopt long-tailed (LT) imbalance by exponentially decreasing the number of samples from the largest to the smallest class. Following~\cite{lee2021abc}, we denote the amount of samples of head class in labeled data and unlabeled data as $N_1$ and $M_1$ respectively. The imbalance ratio for the labeled data and unlabeled data is defined as $\gamma_l$ and $\gamma_u$, which can vary independently. We have $N_k=N_1\cdot{\gamma_l}^{\epsilon_k}$ and $M_k=M_1\cdot{\gamma_u}^{\epsilon_k}$, where $\epsilon_k = \frac{k-1}{K-1}$. 
\textbf{Baseline methods}
For supervised learning, we train network using cross-entropy loss with only labeled data. For semi-supervised learning, we compare the performance of \ourmethod~with FixMatch~\cite{fixmatch}, which do not consider class imbalance. To have a comprehensive comparison, we combine several re-balancing algorithms with FixMatch, including DARP~\cite{darp}, CReST~\cite{wei2021crest}, ABC~\cite{lee2021abc}, DASO~\cite{oh2022daso} and L2AC~\cite{l2ac}.

\textbf{Training and Evaluation} We train Wide ResNet-28-2 (WRN28-2) on CIFAR10-LT, CIFAR100-LT and STL10-LT as a backbone. We evaluate the performance of \ourmethod~using an EMA network, where parameters are updating via exponential moving average every steps, following~\citet{oh2022daso}. We measure the top-1 accuracy on test data and finally report the median of accuracy values of the last 20 epochs following ~\citet{berthelot2019mixmatch}. Each set of experiments was conducted three times. Additional experimental details are provided in the appendix.
\subsection{Results}
\paragraph{In the Case of $\gamma_l = \gamma_u$.}

We initiate our investigation by conducting experiments in the scenario where $\lambda_l = \lambda_u$. 
Our evaluation of the proposed TCBC approach is exhaustive, encompassing a comprehensive comparative analysis against various recent state-of-the-art methods. These methods include DARP~\cite{darp}, CReST+~\cite{wei2021crest}, ABC~\cite{lee2021abc}, DASO~\cite{oh2022daso}, and L2AC~\cite{l2ac}. Further details about these methods are provided in appendix.

The main results on the CIFAR-10 dataset are shown in Table~\ref{table:c10}. It is evident that across various dataset sizes and imbalance ratios, our approach (TCBC) substantially enhances the performance of FixMatch. Moreover, our TCBC consistently surpasses all the compared approaches in these settings, even when they are designed with the assumption of shared class distributions between labeled and unlabeled data. For instance, considering the highly imbalanced scenario with $\gamma_l=\gamma_u=150$, our TCBC  achieves improvements of 2.8\% and 5.0\% in situations where $N_1=1500, M_1=3000$ and $N_1=500, M_1=4000$, respectively, compared to the L2AC.

To facilitate a more comprehensive comparison, we also conducted an evaluation of TCBC using the CIFAR-100 dataset. As illustrated in Table~\ref{table:c100}, our \ourmethod~exhibits a more competitive performance in comparison to the state-of-the-art methods ABC, DASO and L2AC.

\begin{figure*}[!htp]
  \centering
  \includegraphics[width=0.99\textwidth]{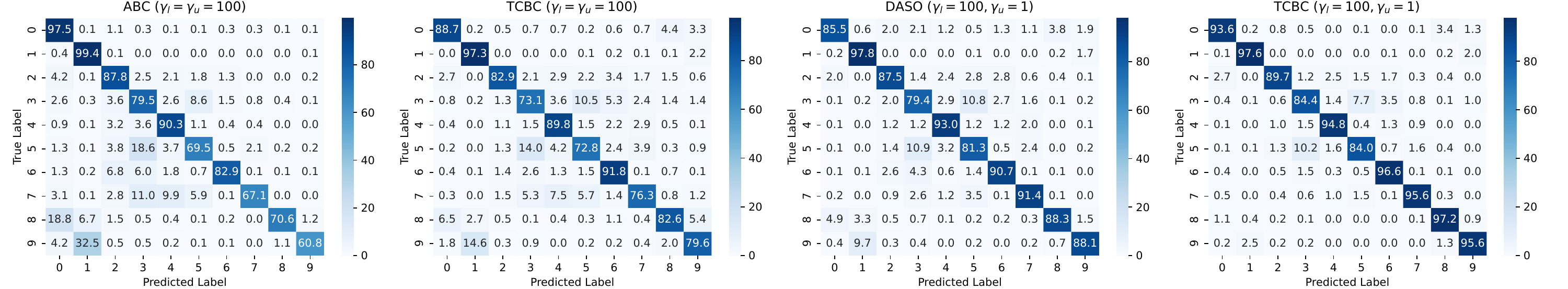}
  \caption{Confusion matrices of ABC, DASO and ours on CIFAR-10 under the imbalance ratio $\gamma_u=100$ and $\gamma_u=1$.}
  \label{fig:cf}
\end{figure*}
\textbf{In the Case of $\gamma_l \neq \gamma_u$.} 
In practical datasets, the distribution of unlabeled data might significantly differs from that of labeled data. Therefore, we explore uniform and reversed class distributions, such as setting $\gamma_u$ to 1 or 1/100 for CIFAR10-LT. In the case of the STL10-LT dataset, as the ground-truth labels of the unlabeled data are unknown, we can only control the imbalance ratio of the labeled data. We present the summarized results in Table~\ref{table:c100}.

Our method demonstrates superior performance when confronted with inconsistent class distributions in unlabeled data. For instance, when $\gamma_{u}$ is set to 1 and 1/100 on CIFAR10-LT, TCBC achieves absolute performance gains of 19.4\% and 17.4\% respectively compared to FixMatch. Similarly, on CIFAR100-LT, our method consistently outperforms compared methods. Even in the case of STL10-LT where the distribution of unlabeled data is unknown, TCBC attains the best results with an average accuracy gain of 3.8\% compared to L2AC. These empirical results across the three datasets with unknown class distributions of unlabeled data validate the effectiveness of TCBC in leveraging unlabeled data to mitigate the negative impact of class imbalance.

\begin{table}[]
\centering
    \caption{Ablation studies of our proposed \ourmethod~ algorithm.}
    \label{}
    \resizebox{\linewidth}{!}{%
\begin{tabular}{@{}lccc@{}}
\toprule
\multicolumn{1}{c}{}       & $\gamma_u=100$ & $\gamma_u=1$ & $\gamma_u=1/100$ \\ \midrule
FixMatch                   & 67.8           & 73.0         & 62.5             \\
w / model bias correction   & 79.2           & 92.3         & 79               \\
w / pseudo-label refinement & 77.8           & 92.1         & 76.8             \\
TCBC                       & 80.3           & 92.4         & 79.9             \\ \bottomrule
\end{tabular}\label{tab:ab}
}
\end{table}
\begin{table}[]
\centering
    \caption{Results on more realistic imbalanced SSL settings.}
\begin{tabular}{@{}ccc|ccc@{}}
\toprule
$\gamma_u$ & DASO                        & \ourmethod       & $\gamma_u$ & DASO & \ourmethod       \\ \midrule
100        & 79.1                        & 84.0(+4.9) & 1/100      & 80.3 & 85.9(+5.6) \\
75         & 80.7                        & 84.3(+3.6) & 1/75       & 82.1 & 86.8(+4.7) \\
50         & 82.9                        & 85.2(+2.3) & 1/50       & 81.6 & 87.6(+6.0) \\
25         &  85.4 & 88.9(+3.5) & 1/25       & 84.0   & 86.0(+2.0)   \\
1          & 88.8                        & 92.8(+4.0) & Avg    & 82.8 & 86.8(+4.0) \\ \bottomrule
\end{tabular} \label{ta:imb_raito}
\end{table}

\subsection{Ablation Study}
To explore the contributions of each key component in TCBC, we conducted a series of ablation studies. We set $N_1$ to 500 and $M_1$ to 4000, and performed experiments on CIFAR-10 with various settings of $\gamma_l=100$. As shown in Table~\ref{tab:ab}, it is evident that using either model bias correction or pseudo-label refinement alone can significantly enhance the performance of FixMatch. This underscores the effectiveness of the two components in our approach.

However, pseudo-label refinement performs poorly when $\gamma_u=100$ and $\gamma_u=1/100$, mainly due to the imbalanced distribution of $\mathcal{D}_l\cup\mathcal{D}_u$, which introduces class bias into the trained model. Model bias correction effectively addresses this issue. Similarly, pseudo-label refinement also enhances the effectiveness of model bias correction.

\subsection{Discussion}
\begin{figure}[t!]
     \centering
     \begin{subfigure}[b]{0.23\textwidth}
         \centering
         \includegraphics[width=\textwidth]{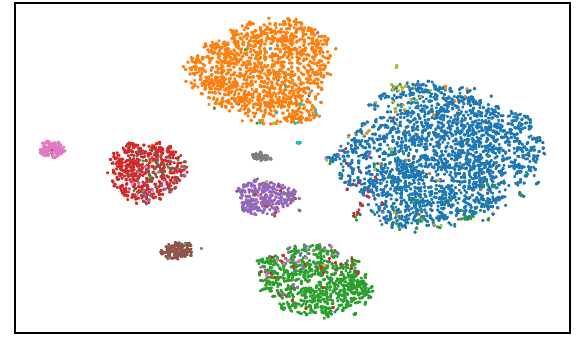}
         \caption{FixMatch ($\gamma_u=100$)}
         \label{fig:2fdsafa}
     \end{subfigure}
     \hfill
     \begin{subfigure}[b]{0.23\textwidth}
         \centering
         \includegraphics[width=\textwidth]{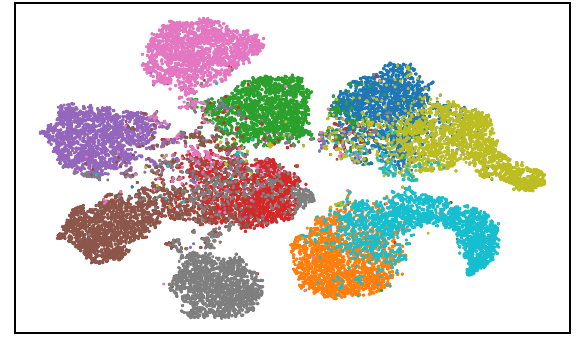}
         \caption{FixMatch ($\gamma_u=1$)}
         \label{fig:2fdfb}
     \end{subfigure}

     \begin{subfigure}[b]{0.23\textwidth}
         \centering
         \includegraphics[width=\textwidth]{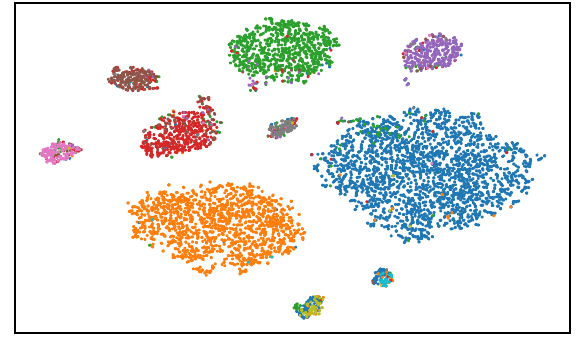}
         \caption{TCBC ($\gamma_u=100$)}
         \label{fig:2fdsafa}
     \end{subfigure}
     \hfill
     \begin{subfigure}[b]{0.23\textwidth}
         \centering
         \includegraphics[width=\textwidth]{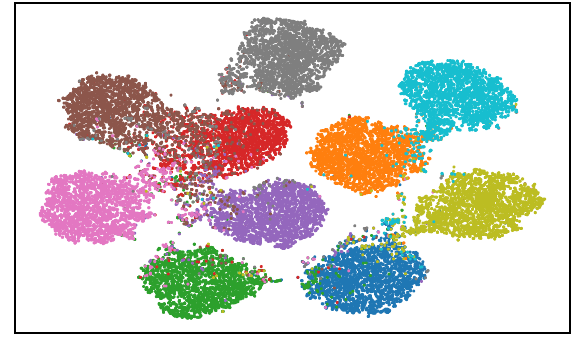}
         \caption{TCBC ($\gamma_u=1$)}
         \label{fig:2fdb}
     \end{subfigure}
        \caption{The t-SNE visualization of the unlabeled set on CIFAR-10-LT dataset.}
        \label{fig:cfm}
\end{figure}
\textbf{Can our method adapt to unlabeled samples with different class distributions?} To assess the effectiveness of our approach across different distributions of unlabeled samples, supplementary experiments were conducted using the CIFAR-10-LT dataset. The labeled sample distribution $\gamma_l$ was maintained at a constant value of 100, while systematically modifying the imbalance ratio ($\gamma_u$) of unlabeled data. In this study, we configured $N_1$ as 1500 and $M_1$ as 3000, and we conducted a comparative analysis of our method against the performance of DASO. The outcomes of the experiments are illustrated in Table \ref{ta:imb_raito}. Notably, our TCBC method consistently demonstrated superior performance compared to DASO across all test scenarios, achieving an average performance increase of 4\%. 
These results clearly demonstrate that our method can effectively adapt to imbalanced SSL environments in the real world.

\textbf{How does our method perform on samples of different frequencies?} 
A comparative analysis was performed on our TCBC in comparison to various configurations of ABC and DASO. Figure~\ref{fig:cf} shows the confusion matrices of the models obtained by different algorithms on the test set when $N_1$ is set to 1500 and $M_1$ is set to 3000. The two confusion matrices on the left depict the test results of ABC and TCBC when the labeled and unlabeled class distributions are consistent. From the recall of the minority class samples, it is evident that ABC exhibits bias across different classes. However, our approach effectively mitigates such biases. Under the settings of $\gamma_l=100$ and $\gamma_u=1$, both our method and the model learned by DASO reduced the model's class bias. However, our method outperforms DASO on both majority and minority classes. This observation indicates that our TCBC has successfully learned a model without class bias.

\begin{figure}
     \centering
     \begin{subfigure}[b]{0.23\textwidth}
         \centering
         \includegraphics[width=\textwidth]{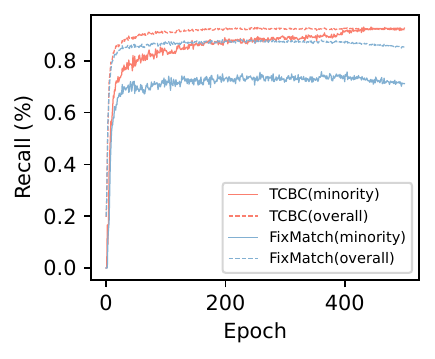}
         \caption{$\gamma_l=\gamma_u = 100 $}
         \label{figfdsaf:2a}
     \end{subfigure}
     \hfill
     \begin{subfigure}[b]{0.235\textwidth}
         \centering
         \includegraphics[width=\textwidth]{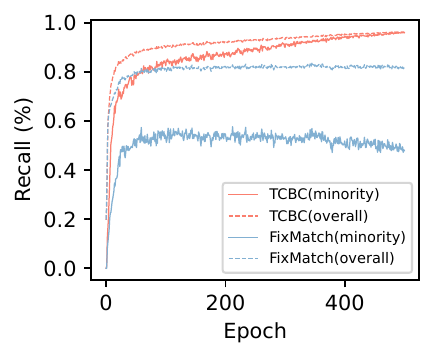}
         \caption{$\gamma_l=100, \gamma_u = 1$}
         \label{fifdsafg:2b}
     \end{subfigure}
        \caption{Train curves for the recall of pseudo-labels.}
        \label{fig:rccc}
\end{figure}

\textbf{Have our methods learned better features?}
We further visualized the features of unlabeled samples using t-SNE~\cite{van2008visualizing} in the same setting as depicted in Figure~\ref{fig:cf}. As shown in Figure \ref{fig:cfm}, compared to FixMatch, TCBC has learned more discriminative features. For instance, in the case of $\gamma_u=100$, FixMatch exhibits only 8 clusters, while our method demonstrates all 10 clusters. This high-quality feature representation reflects the effectiveness of our method in learning an unbiased model and refining pseudo-labels.

\textbf{How does the TCBC enhance model performance?}
 We examined the changes in recall of minority and all classes of unlabeled samples under the settings of Figure~\ref{fig:cf} for both FixMatch and TCBC methods. As shown in Figure~\ref{fig:rccc}, it is evident that in FixMatch, the recall of the minority class quickly plateaus and exhibits a significant disparity from the recall of all classes. In contrast, in TCBC, the recall of the minority class continues to increase and approaches the average recall. This indicates that TCBC ensures equitable treatment of pseudo-labeling for minority classes throughout the process, aligning with our motivation.

\section{Conclusion}
In this work, we address the model bias and pseudo-label bias in imbalanced SSL through the introduction of a novel twice correction approach. For tackling model bias, we propose the utilization of an estimate of the training sample class distribution to rectify the model's learning objectives onto an unbiased posterior probability. To address pseudo-label bias, we refine a better set of pseudo-labels by estimating the class bias under the current parameters during the training process. Extensive experimental results demonstrate that our method outperforms existing approaches.

\section{Acknowledgments}
This work is supported by the National Science
Foundation of China (61921006). We would like
to thank Xin-chun Li, and the anonymous reviewers for their helpful discussions and support. 

\bibliography{aaai24}

\end{document}